\documentclass[12pt]{article}

\usepackage[]{graphicx}
\usepackage[]{color}
\usepackage{alltt}
\usepackage{lipsum} 
\usepackage{amsmath}
\usepackage{wrapfig}
\usepackage{subcaption}
\DeclareMathOperator*{\argmax}{arg\,max}

\usepackage{parskip}
\newcommand{\mytitle}{Introduction and Exemplars of Uncertainty Decomposition}
\newcommand{\myname}{Shuo Chen}
\newcommand{\mysupervisor}{Dr. David Rügamer, Chris Kolb}

\usepackage[a4paper, width = 160mm, top = 35mm, bottom = 30mm, 
bindingoffset = 0mm]{geometry}
\usepackage[utf8]{inputenc}
\usepackage{ragged2e}
\usepackage{xcolor}
\usepackage{amsfonts}

\usepackage[round, comma]{natbib}
\usepackage{fancyhdr}
\newcommand{\changefont}{%
    \fontsize{8}{11}\selectfont
}
\usepackage{hyperref}
\hypersetup{
  colorlinks = true,
  linkcolor = black,
  urlcolor = black,
  citecolor = black}
\IfFileExists{upquote.sty}{\usepackage{upquote}}{}
\begin{document}

 
\begin{titlepage}
\begin{center}
    
\LARGE
UQDL Seminar Report 
    
\vspace{0.5cm}
      
\rule{\textwidth}{1.5pt}
\LARGE
\textbf{\mytitle}
\rule{\textwidth}{1.5pt}
   
\vspace{0.5cm}
      
\large
Department of Statistics \\
Ludwig-Maximilians-Universität München 

\vfill

\Large
\textbf{\myname}

\vfill

\large
Munich, 1 13\textsuperscript{th}, 2022
      
\vfill

\includegraphics[width = 0.4\textwidth]{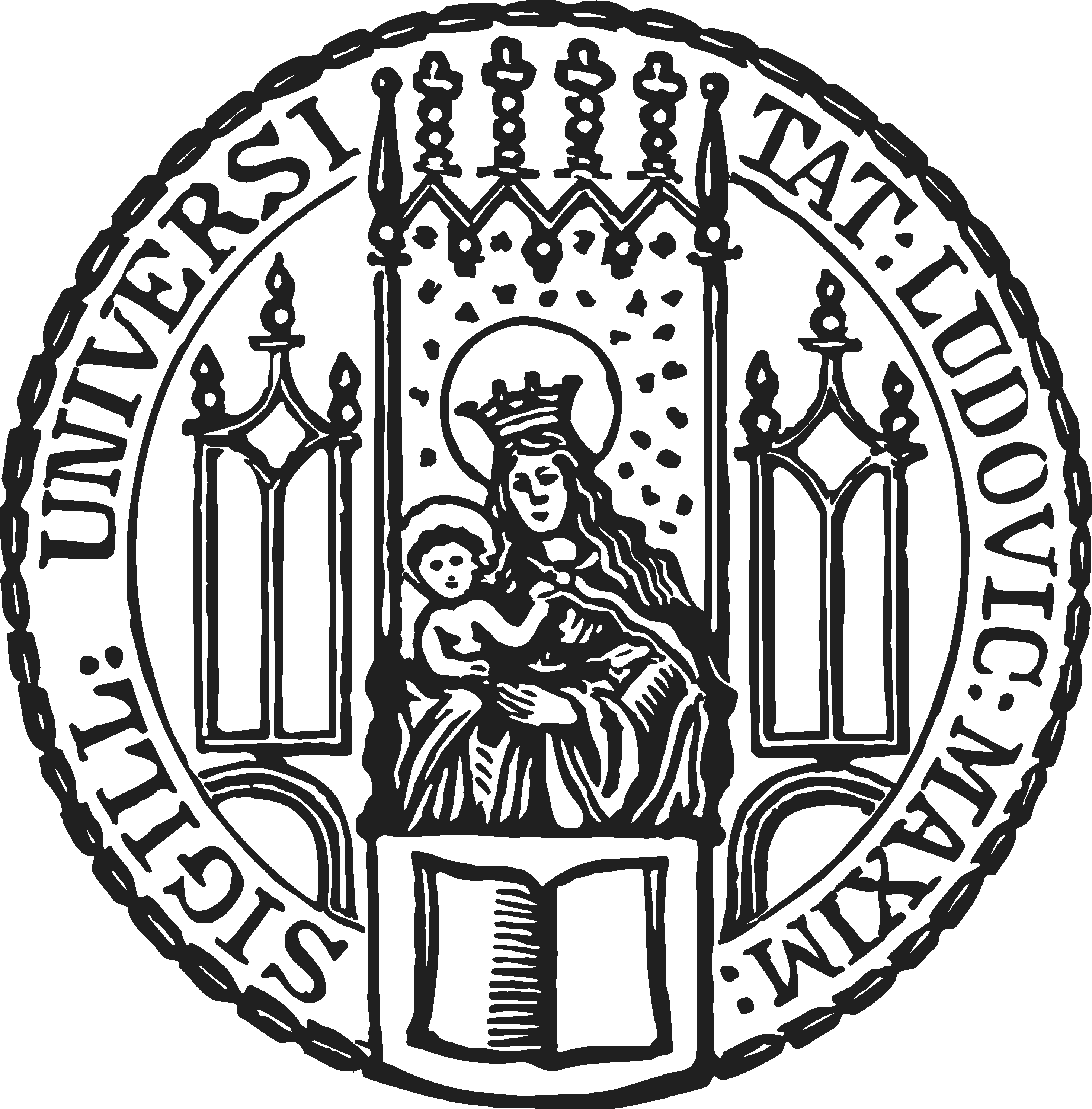}

\vfill

\normalsize
Report for the seminar Uncertainty Quantification in Deep Learning
\\

Supervised by \mysupervisor

\end{center}
\end{titlepage}


\pagenumbering{Roman}
\newpage

\begin{abstract}

Uncertainty plays a crucial role in the machine learning field. Both model trustworthiness and performance require the understanding of uncertainty, especially for models used in high-stake applications where errors can cause cataclysmic consequences, such as medical diagnosis and autonomous driving. Accordingly, uncertainty decomposition and quantification have attracted more and more attention in recent years. This short report aims to demystify the notion of uncertainty decomposition through an introduction to two types of uncertainty and several decomposition exemplars, including maximum likelihood estimation, Gaussian processes, deep neural network, and ensemble learning. In the end, cross connections to other topics in this seminar and two conclusions are provided.

\end{abstract}

\newpage
\tableofcontents

\newpage


\pagenumbering{arabic}

\section{Introduction}
\label{intro}

\subsection{The Necessity of Uncertainty Quantification}

In the past decades, tremendous success across a variety of fields has shown the enormous potential of machine learning and deep learning algorithms \citep{lecun2015deep}. Therefore, the predictive results of these models are being used in more and more critical applications in which a minor error could lead to catastrophic consequences, such as medical diagnoses \citep{begoli2019need}, drug design \citep{begoli2019need}, self-driving cars \citep{DBLP:journals/corr/KendallG17}. 

\begin{figure}[h]

\begin{subfigure}{0.6\textwidth}
\includegraphics[width=1\linewidth]{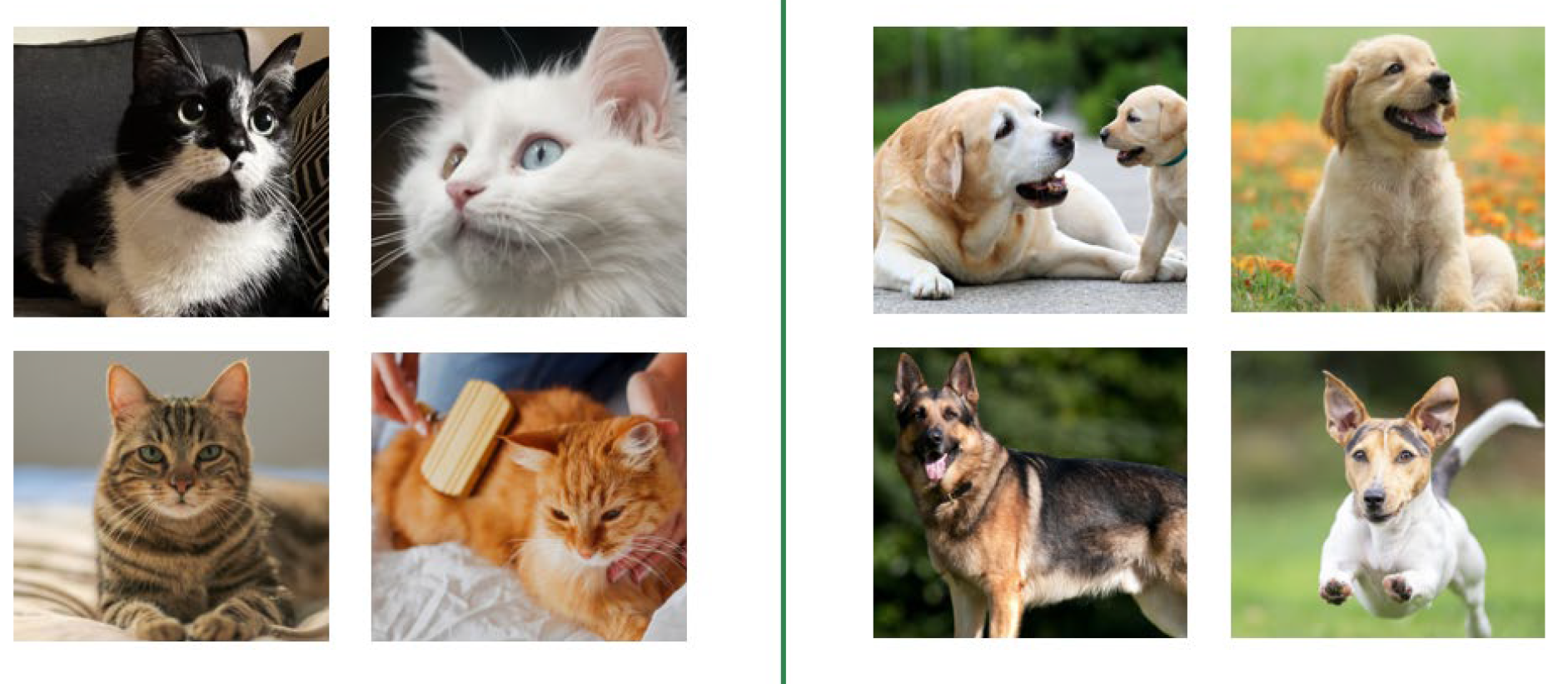} 
\caption{An animal classification task.}
\label{fig:catanddog}
\end{subfigure}
\begin{subfigure}{0.4\textwidth}
\includegraphics[width=1\linewidth]{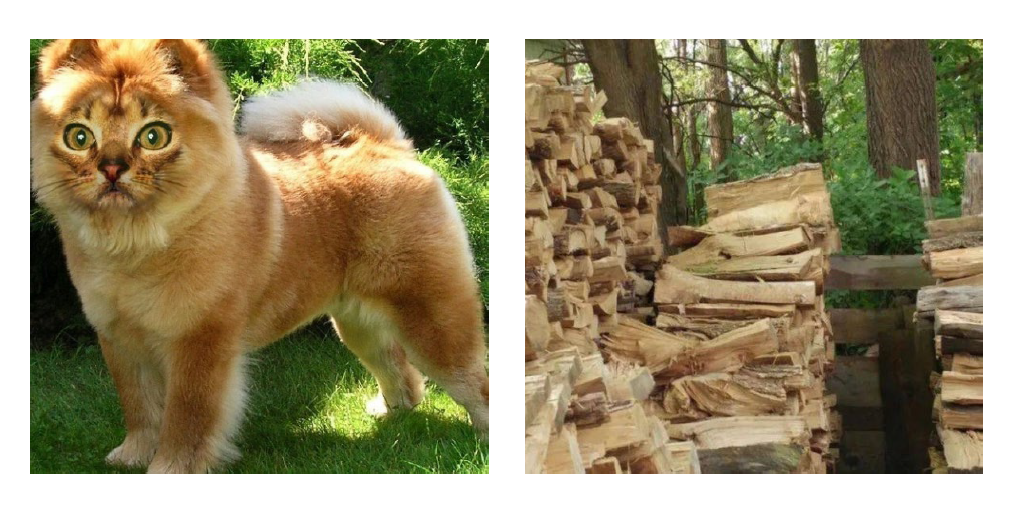}
\caption{Two uncertain animals.}
\label{fig:catordog}
\end{subfigure}

\caption{Animal classification with uncertain examples.}
\label{fig:image2}
\end{figure}

As machine learning aims to train a model from given data and predict new observations using the trained model, it is inherently connected with the notion of uncertainty \citep{hullermeier2021aleatoric}. Uncertainty is ubiquitous in our world and natural to humans. For instance, if we look at Figure \ref{fig:catanddog} and Figure \ref{fig:catordog} to specify the animals shown in the pictures. Statements like "I am pretty sure there are dogs on the right and cats on the left in Figure \ref{fig:catanddog}." and "I am not sure whether these pictures in Figure \ref{fig:catordog} are cats or dogs." are likely to emerge. 



Machine learning algorithms should also have the ability to specify the uncertainty they encounter during training and testing, especially in those high-stake applications mentioned before where we need to make sure that our algorithms do not make mistakes by any chance. But we cannot expect them to be a panacea all the time. However, we could train them to inform us if they are uncertain about the task. In this way, double-check from a human specialist could be incorporated as well. In other words, point predictions and accuracy are not enough for training and evaluating these models. Accurate uncertainty quantification (UQ) of predictive uncertainty such as meaningful confidence intervals is also required \citep{ovadia2019can}. 

In the following content, two main types of uncertainty are introduced in Section \ref{section 1.2}. In Section \ref{decomposition}, there are several examples of uncertainty decomposition in 4 famous machine learning algorithms, i.e., maximum likelihood estimation, Gaussian processes, deep neural network, and ensemble learning. Section \ref{connections} shows cross connections with other topics in this seminar. In the end, limitations of this report are discussed and two conclusions are provided in Section \ref{conclusion}.

\subsection {Uncertainty Taxonomy}
\label{section 1.2}

\begin{figure}[ht!]
\centering
\includegraphics[width=0.7\textwidth]{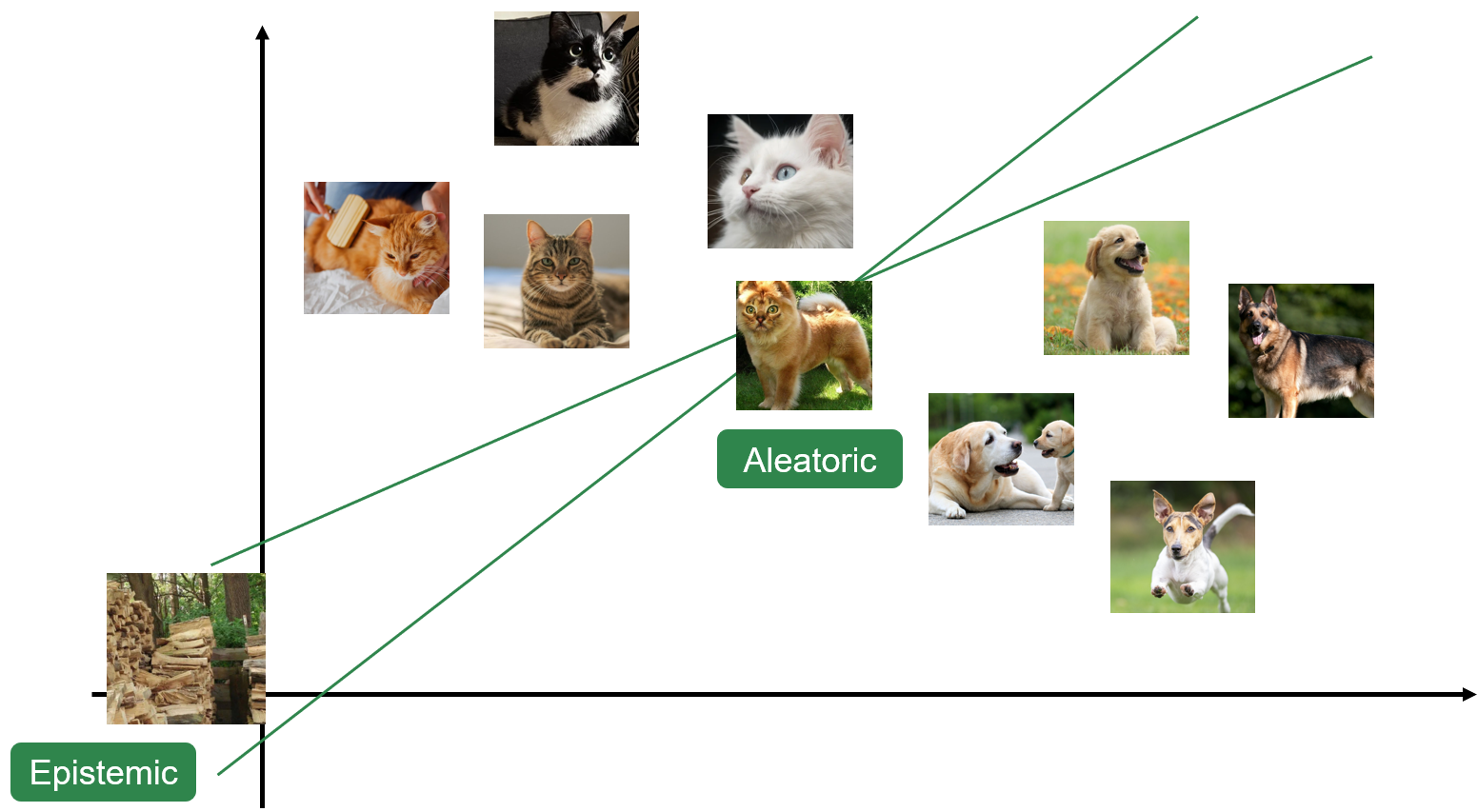}
\caption{Animal pictures in an assumed two-dimensional feature space.}
\label{fig:features}
\end{figure}

Let us first have another look at Figure \ref{fig:catanddog} and \ref{fig:catordog} and suppose that we could place all these pictures in a two-dimensional feature space based on their similarity as shown in Figure \ref{fig:features}. We will notice that the two uncertain pictures are difficult to recognize due to different reasons. For one, it is because the features of cats and dogs are mixed together. Hence, the picture lies close both to other cats' and dogs' pictures. For the other, the problem is that such a picture is rare and the features from animals are hard to recognize which makes it lie far away from other pictures. These two bizarre examples illustrate the two main types of uncertainty in machine learning which are conceptually called \textit{aleatoric} and \textit{epistemic uncertainty}. 

Aleatoric uncertainty is also known as \textit{data uncertainty} which is an inherent property of the data generation process. In most of the cases, we assume that there is a true but unknown probability measures $P_{\theta^*}$ with parameter $\theta^*$ generating the data. Given such an assumption, this process always has a stochastic component that cannot be reduced by any additional source of information \citep{hullermeier2021aleatoric}. Coin-flipping could serve as a prototypical example. No matter how accurate our prediction function is, it could only give probabilities for the two possible outcomes instead of a definite answer. Due to such inherently random effects, aleatoric uncertainty is irreducible. 

Epistemic uncertainty, on the other hand, is reducible and also known as \textit{knowledge uncertainty}. It refers to uncertainty originated from a lack of knowledge about the best model and could be explained away given enough data \citep{DBLP:journals/corr/KendallG17}. In other words, it stands for the ignorance of our model and for the epistemic state of the model itself rather than any other underlying random phenomenon \citep{hullermeier2021aleatoric}. For example, if a weather forecaster says that he or she is very certain that the chance of rain is 50\%, here the 50\% means aleatoric uncertainty. But a statement like "the best estimate at 30\% is very uncertain due to lack of weather data" is more akin to epistemic uncertainty \citep{Kull20114} where if given more data the forecaster could be more confident about the estimation. 



\begin{figure}[h]

\begin{subfigure}{0.4\textwidth}
\includegraphics[width=1\linewidth]{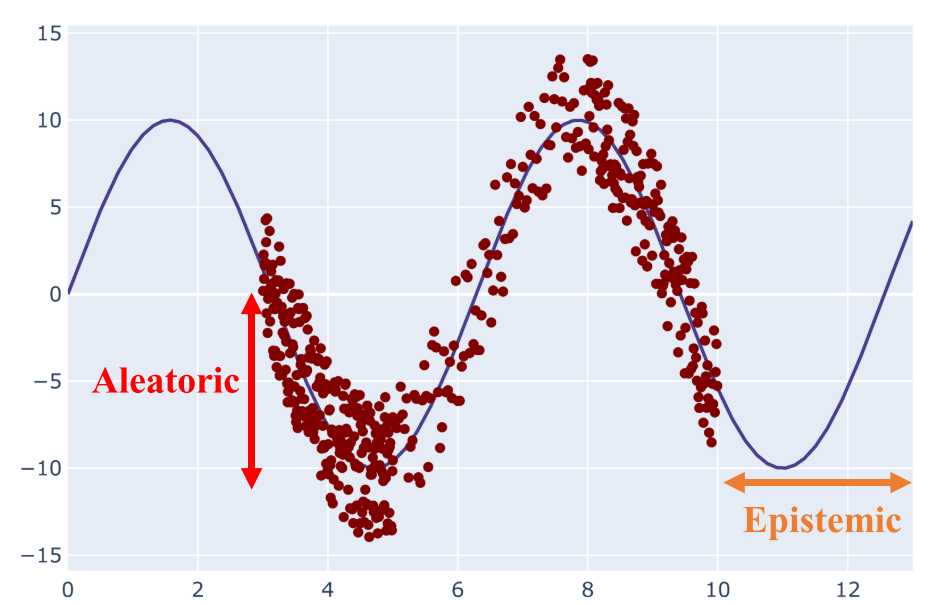} 
\caption{Two uncertainties in regression.}
\label{fig:regression}
\end{subfigure}
\begin{subfigure}{0.6\textwidth}
\includegraphics[width=1\linewidth]{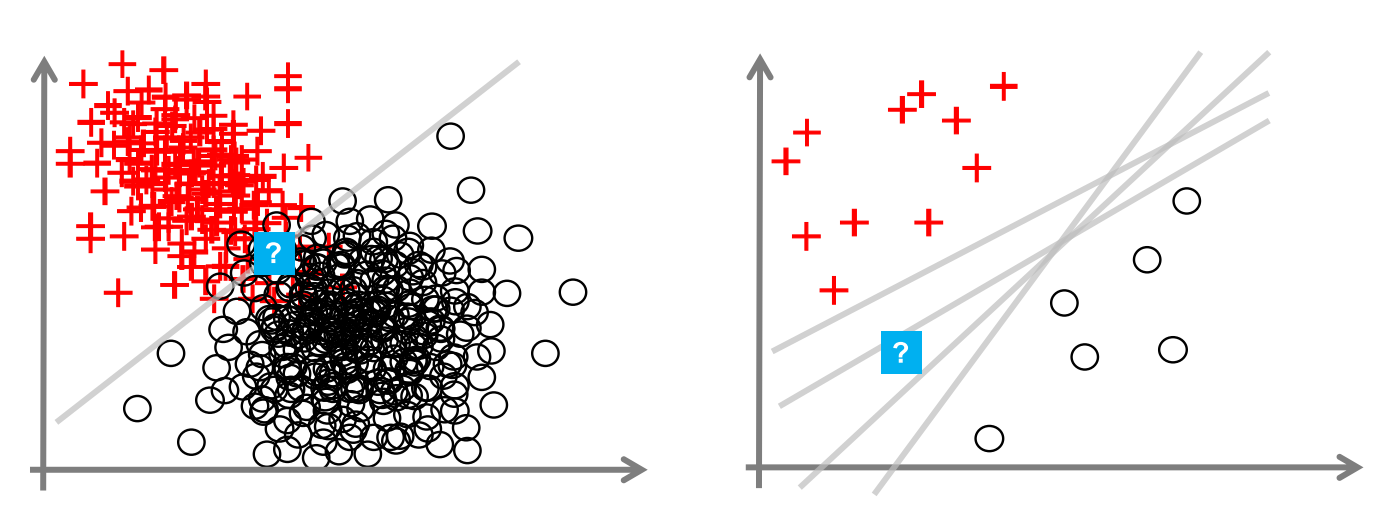}
\caption{Two uncertainties in classification.}
\label{fig:classification}
\end{subfigure}

\caption{An illustration of two uncertainties in machine learning tasks.}
\label{fig:image2}
\end{figure}

Figure \ref{fig:regression} \citep{abdar2021review} illustrates these two different uncertainties in a regression setting. In the left region, training data overlap which makes the prediction uncertain. Hence there is a high aleatoric uncertainty. While on the right, there is no data available and the model suffers from the lack of knowledge from data which increases the epistemic uncertainty. Similarly, Figure \ref{fig:classification} \citep{hullermeier2021aleatoric} visualizes these uncertainties in a classification setting. On the left, data points with opposite labels mix together which makes the aleatoric uncertainty higher. But on the right, there is no training data available in the region which intensifies epistemic uncertainty. 

\begin{figure}[ht!]
\centering
\includegraphics[width=0.5\textwidth]{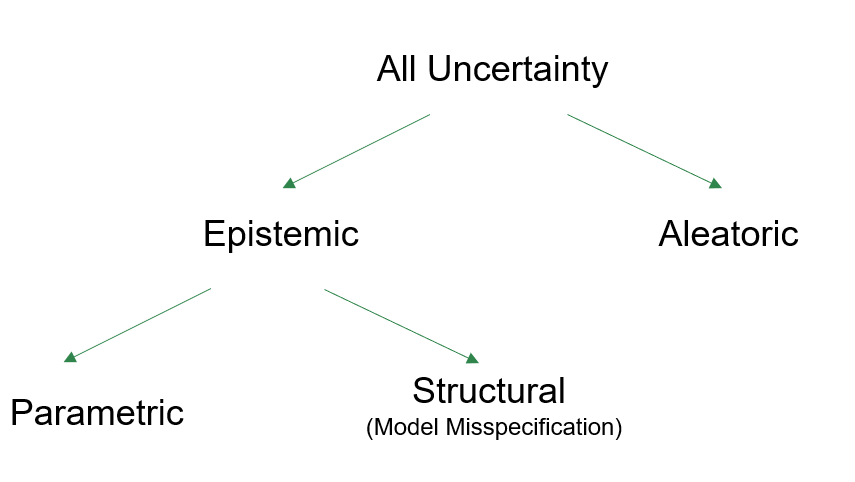}
\caption{A decomposition of different types of uncertainty.}
\label{fig:branch}
\end{figure}

Furthermore, a machine learning model's epistemic uncertainty can emerge from two sources \citep{sullivan2015introduction}: \textit{parametric uncertainty} and \textit{structural uncertainty} (see Figure \ref{fig:branch}). The first one emphasize uncertainty related to model parameter estimations under current model specification and the second one reflects the discrepancy between the assumed model specification and the true but unknown data-generating process. In some literature, they are also known as \textit{approximation uncertainty} and \textit{model uncertainty} \citep{hullermeier2021aleatoric}.

\section{Uncertainty Decomposition Exemplars}
\label{decomposition}

Aleatoric uncertainty and epistemic uncertainty originate from different sources, vary in many aspects, and therefore cannot be treated the same. The goal of uncertainty quantification is to properly characterize both aleatoric and epistemic uncertainties \citep{DBLP:journals/corr/abs-1911-04061}. In areas that training data have adequately represented, aleatoric uncertainty should precisely estimate the data-generation distribution. On the other hand, in areas lacking training data, epistemic uncertainty should increase to indicate a lack of confidence in predictions. Under such circumstances, an accurate decomposition of the whole uncertainty is preferred. 

This section demonstrates exemplars of uncertainty decomposition in four well-known and essential machine algorithms, including maximum likelihood estimation, Gaussian processes, deep neural networks, and ensemble learning. In each sub-section, a brief recap of each algorithm is presented, and an explanation of uncertainty decomposition follows.

\subsection{Maximum Likelihood Estimation}
Likelihood is a key element of statistical inference and maximum likelihood estimation serves as a workhorse in machine learning. Many learning algorithms including those for training deep neural networks achieve model induction as likelihood maximization  \citep{hullermeier2021aleatoric}. 

Considering a data-generating process parametrized by probability measures $P_{\boldsymbol{\theta}^*}$ where $\boldsymbol{\theta}^* \in \Theta$ stands for the true but unknown parameter vector. An essential problem is to estimate $\boldsymbol{\theta}^*$, i.e., to identify the underlying data-generating process based on a set of observations $\mathcal{D}=\{X_1,\dots,X_N\}$ generated by $P_{\boldsymbol{\theta}^*}$. Maximum likelihood estimation (MLE) is one method to approach such problem. The core idea of MLE is to estimate $\boldsymbol{\theta}^*$ by maximizing the likelihood function, or equivalently, the log-likelihood function. Assuming that observations in $\mathcal{D}=\{X_1,\dots,X_N\}$ are independent and $f(\cdot)$ is the density function of $P_{\boldsymbol{\theta}^*}$, the log-likelihood function is 
\begin{equation}
    \ell_{N}(\boldsymbol{\theta}):=\sum_{n=1}^{N} \log f\left(X_{n}\right),
\end{equation}
and the corresponding result from MLE is 
\begin{equation}
    \boldsymbol{\theta}^*_f = \argmax_{\boldsymbol{\theta}} \ell_N(\boldsymbol{\theta}) 
    = \argmax_{\boldsymbol{\theta}} \sum_{n=1}^{N} \log f\left(X_{n}\right).
\end{equation}

Two prototypical indicators describe the uncertainty in MLE, one is \textit{Fisher information matrix}  and the other is \textit{Akaike Information Criteria} (AIC).

\begin{figure}[ht!]
\centering
\includegraphics[width=0.7\textwidth]{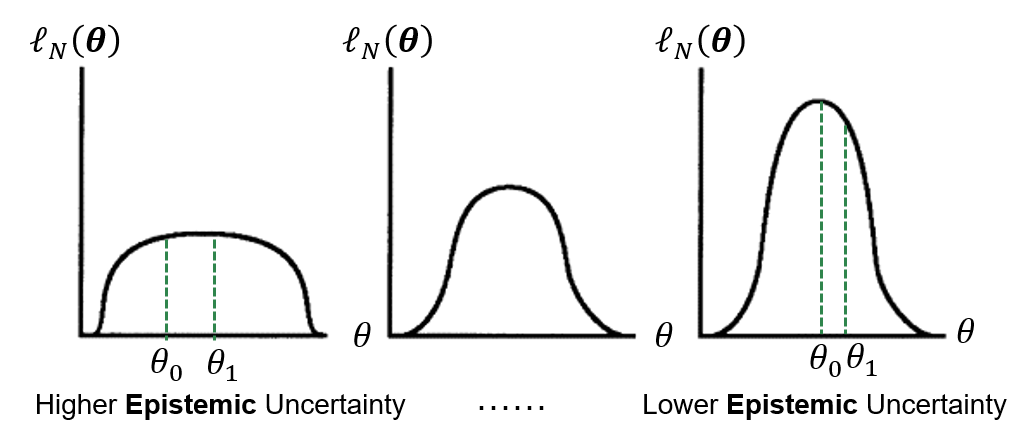}
\caption{Log-likelihood function with different peakiness.}
\label{fig:mle}
\end{figure}

Fisher information matrix is the negative Hessian of the log-likelihood function and formally written as
\begin{equation}
 \mathcal{I}_{\mathcal{N}}(\boldsymbol{\theta})=-\left[\mathbf{E}_{\boldsymbol{\theta}}\left(\frac{\partial^{2} \ell_{N}}{\partial \theta_{i} \partial \theta_{j}}\right)\right]_{1 \leq i, j \leq N}.   
\end{equation}
It is the expected second-order derivative of the likelihood function, and it quantifies the curvature of the likelihood. In particular, at the maximum of the likelihood function, where the slope is zero, the second-order derivative expresses the "peakiness" of the maximum \citep{kauermannstatistical}. If the likelihood function is peaked, the model is more certain to specify the parameter, so there is less epistemic uncertainty. Consider the graphs in Figure \ref{fig:mle}, the first slope is gentle and a bunch of estimates have similar likelihood, so we are uncertain which one is better. But the last is steep, and we could be more confident to specify the estimate. Furthermore, it allows for constructing confidence regions for the target parameter $\boldsymbol{\theta}^*$ around the estimate $\boldsymbol{\theta}^*_f$. The larger the region, the higher the epistemic uncertainty about the true model \citep{hullermeier2021aleatoric}. 

Recall that we assume the density function of $P_{\boldsymbol{\theta}^*}$ is $f(\cdot)$ and derive the whole estimation afterwards, however, we actually don't know the true form of $P_{\boldsymbol{\theta}^*}$. Under such circumstance, AIC is a good tool to measure the distance from the true parameter $\boldsymbol{\theta}^*$ and MLE result $\boldsymbol{\theta}^*_f$. Formally speaking, AIC is defined as 
\begin{equation}
    A I C=-2 \sum_{i=1}^{N} \log f\left(X_{i} ; \boldsymbol{\theta}^*_f\right)+2 p,
\end{equation}
where $p$ is the dimension of the parameter vector. It is an approximation of the second term of the expected Kullback-Leibler Divergence which indicates the distance between the true density function of $P_{\boldsymbol{\theta}^*}$ and  our assumption $f(\cdot)$ \citep{kauermannstatistical}. 

\begin{figure}[ht!]
\centering
\includegraphics[width=0.9\textwidth]{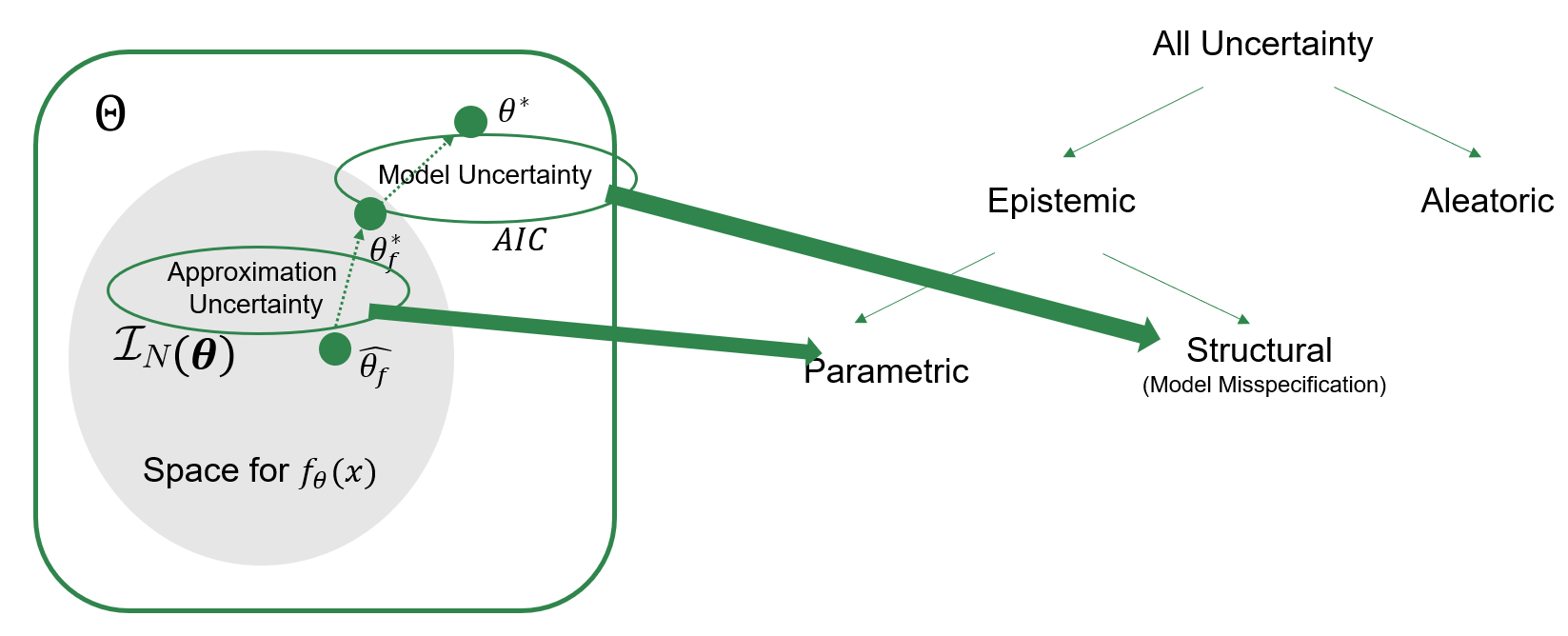}
\caption{Relationship between AIC and Fisher information matrix.}
\label{fig:mle2}
\end{figure}

A more detailed illustration of the relationship between AIC, Fisher information matrix, and uncertainty is shown in Figure \ref{fig:mle2}. The difference between the best guess $\boldsymbol{\theta}^*_f$ given by MLE  within our hypothesis space and the ground truth $\boldsymbol{\theta}^*$ is evaluated by AIC. It is also the structural uncertainty or model uncertainty which is a branch of epistemic uncertainty and caused by model misspecification. 

However, due to limited observations and computation resources, even $\boldsymbol{\theta}^*_f$ cannot be prescribed, the discrepancy between our actual result $\hat{\boldsymbol{\theta}}_f$ and the theoretically best estimate $\boldsymbol{\theta}^*_f$ could be evaluated by Fisher information matrix which also indicates the approximation uncertainty or parametric uncertainty.

\subsection{Gaussian Processes}
\label{gaussian}
Gaussian processes have been widely used in machine learning applications due to its representation flexibility and inherently uncertainty measures over predictions \citep{wang2021intuitive}. A Gaussian process model is a probability distribution over possible functions that fit a set of points and any finite sample of functions are jointly Gaussian distributed. As it is essentially a probability distribution, we can calculate its mean and variances to indicate how confident the model is. 

More specifically, a collection of random variables $\{f(\mathbf{x})|\mathbf{x}\in \mathcal{X}\}$ with index set $\mathcal{X}$ is said to be drawn from a Gaussian process with mean function $m$ and covariance function $k$, if for any finite set of elements $\mathbf{x}_1, \dots, \mathbf{x}_n \in \mathcal{X}$, the set of random variables $f(\mathbf{x}_1),\dots,f(\mathbf{x}_n)$ follows multivariate normal distribution 

\begin{equation}
    \left[\begin{array}{c}
f\left(\mathbf{x}_{1}\right) \\
f\left(\mathbf{x}_{2}\right) \\
\vdots \\
f\left(\mathbf{x}_{n}\right)
\end{array}\right] \sim \mathcal{N}\left(\left[\begin{array}{c}
m\left(\mathbf{x}_{1}\right) \\
m\left(\mathbf{x}_{2}\right) \\
\vdots \\
m\left(\mathbf{x}_{n}\right)
\end{array}\right],\left[\begin{array}{ccc}
k\left(\mathbf{x}_{1}, \mathbf{x}_{1}\right) & \cdots & k\left(\mathbf{x}_{1}, \mathbf{x}_{n}\right) \\
\vdots & \ddots & \vdots \\
k\left(\mathbf{x}_{n}, \mathbf{x}_{1}\right) & \cdots & k\left(\mathbf{x}_{n}, \mathbf{x}_{n}\right)
\end{array}\right]\right).
\end{equation}
Such process is denoted as $f \sim \mathcal{GP}(m,k)$ where 
\begin{equation}
    \begin{aligned}
m(\mathbf{x}) &=\mathbf{E}(f(\mathbf{x})) ,\\
k\left(\mathbf{x}, \mathbf{x}^{\prime}\right) &=\mathbf{E}\left((f(\mathbf{x})-m(\mathbf{x}))\left(f\left(\mathbf{x}^{\prime}\right)-m\left(\mathbf{x}^{\prime}\right)\right)\right).
\end{aligned}
\end{equation}

In other words, a function $f$ drawn from a Gaussian process can be treated as a high-dimensional vector drawn from a high-dimensional multivariate Gaussian distribution \citep{hullermeier2021aleatoric}. 

Inference from Gaussian processes model follows the Bayesian inference paradigm. Starting from a prior distribution specified by a mean function $m$ and kernel $k$, then this prior distribution could be replaced by a posterior on the basis of observations $\mathcal{D}=\{(\boldsymbol{x_i},y_i)\}_{i=1}^N$, where we assume each output $y_i=f(\mathbf{x}_i)+\epsilon_i$ has an additional noise component $\epsilon_i$. After imposing a Gaussian noise with variance $\sigma^2_{\epsilon}$ and a zero-mean prior,that is to say $y_i$ is normally distributed with expected value $f(\boldsymbol{x_i})$ and variance $\sigma^2_{\epsilon}$, the posterior predictive distribution for a new query $\mathbf{x}_q$ is again a Gaussian distribution with mean $\mu$ and variance $\sigma^2$ as follows:
\begin{equation}\label{eq:7}
\begin{aligned}
\mu &=K\left(\mathbf{x}_{q}, X\right)\left(K(X, X)+\sigma_{\epsilon}^{2} I\right)^{-1} \boldsymbol{y}, \\
\sigma^{2} &=K\left(\mathbf{x}_{q}, \mathbf{x}_{q}\right)+\sigma_{\epsilon}^{2}-K\left(\mathbf{x}_{q}, X\right)\left(K(X, X)+\sigma_{\epsilon}^{2} I\right)^{-1} K\left(X, \mathbf{x}_{q}\right),
\end{aligned}
\end{equation}
where $X$ is an $N \times d$ matrix containing training data, $K(X,X)$ is the kernel matrix with entries $(K(X,X))_{i,j} = k(\mathbf{x}_i,\mathbf{x}_j)$ and $\boldsymbol{y}$ is the vector of observed training targets. 

Moreover, Gaussian processes can also approach classification problems where observations have discrete targets by extra link functions such as logistic link function. However, under such circumstance, the posterior predictive distribution is no longer Gaussian, approximation inference techniques such as Laplace, MCMC will be required \citep{hullermeier2021aleatoric}. 

The posterior predictive distribution given in equation \ref{eq:7} naturally captures the uncertainty. Its variance $\sigma^2$ for a query $\mathbf{x}_q$ describes the total uncertainty while making the prediction $y_q$. The higher $\sigma^2$, the more uncertain the model and the less trustworthy the result. Besides, to describe the inherent stochastic component of the data generation process, we pose an additional noise term followed a Gaussian distribution with variance $\sigma^2_{\epsilon}$. Hence $\sigma^2_{\epsilon}$ exactly corresponds to the aleatoric uncertainty. Accordingly, the difference between $\sigma^2$ and $\sigma^2_{\epsilon}$ could be treated as epistemic uncertainty. 

\begin{figure}[ht!]
\centering
\includegraphics[width=1\textwidth]{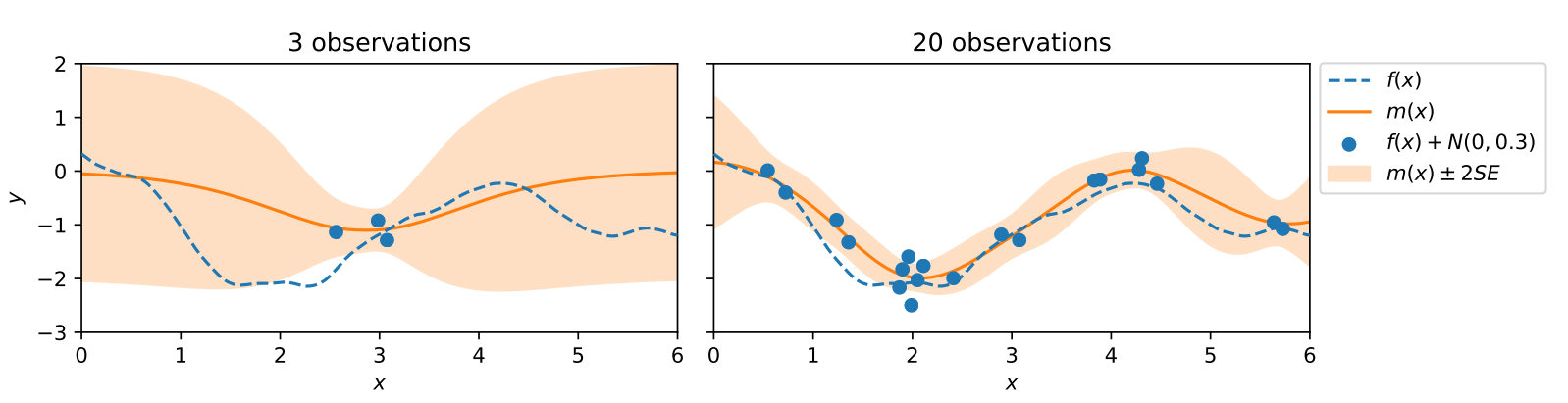}
\caption{An one-dimensional illustration of Gaussian processes.}
\label{fig:gp}
\end{figure}

Figure \ref{fig:gp} \citep{hullermeier2021aleatoric} illustrates a simple one-dimensional regression setting with very few observations on the left and more on the right. The whole predictive uncertainty is reflected by the width of the orange variance band around the mean function. With an increasing number of observations, the uncertainty reduces to some extent.

\subsection{Deep Neural Networks}
\label{bnn}
Deep learning methods have drastically improved state-of-the-art in speech recognition, visual object recognition, object detection and many other domains such as drug discovery and genomics \citep{lecun2015deep}. \textit{Deep neural networks}, also often called \textit{deep feedforward networks} or \textit{multilayer perceptrons} (MLPs), are the quintessential deep learning methods \citep{goodfellow2016deep}. 


Extensions of feedforward neural networks such as \textit{deep convolution neural networks} \citep{krizhevsky2012imagenet}, \textit{recurrent neural networks} \citep{hochreiter1997long}, \textit{attention mechanisms} \citep{vaswani2017attention} have shown great promises in many domains.


Generally speaking, all of these architectures can be seen as a probabilistic classifier $f$ with parameters $\boldsymbol{\theta}$. For classification , given a query $\mathbf{x}$, the final layer typically outputs a probability distribution on the set of classes $\mathcal{C}$. Training of these models is essentially doing maximum likelihood estimation by back-propagation and minimizing the corresponding loss.

Although deep neural networks hold great promise, they failed to report the confidence about their predictions. For instance, the probability output in Figure \ref{fig:cnn} is not an indicator of the model's confidence. The reason could be that during training, the dataset simply doesn't contain information about the uncertainty for each observation, as our labels are all deterministic, and the model can learn to represent the uncertainty from nowhere. In other words, their probability predictions only captures aleatoric uncertainty but epistemic uncertainty is ignored \citep{hullermeier2021aleatoric}. 

To indicate epistemic uncertainty, \textit{Bayesian neural networks} (BNNs) have been proposed as a Bayesian extension of deep neural networks \citep{neal2012bayesian}. In BNNs, each parameter is no longer a real number but represented by a probability distribution, which is usually Gaussian distribution; The learning process turns to be Bayesian inference. Formally, we assume parameters $\boldsymbol{\theta} \sim P_{\gamma}$ follow a prior distribution parameterized by $\gamma$. Given a dataset $\mathcal{D}$, we compute the posterior distribution $p(\boldsymbol{\theta}|\mathcal{D})$. When a new query $\mathbf{x}_q$ comes, the prediction $y$ is given by the predictive posterior distribution 
\begin{equation}
p\left(y \mid \mathbf{x}_{q}, \mathcal{D}\right)=\int p\left(y \mid \mathbf{x}_{q}, \boldsymbol{\boldsymbol{\theta}}\right) p(\boldsymbol{\theta} \mid \mathcal{D}) d \boldsymbol{\theta},
\end{equation}
which could be further used to represent the epistemic uncertainty. 

\cite{depeweg2018decomposition} have proposed a promising manner for measuring and separating aleatoric and epistemic uncertainty based on the predictive posterior distribution. They chose the \textit{entropy} of the predictive distribution, that is, 

\begin{equation}
H[p(y \mid \mathbf{x}_q)]=-\sum_{y \in \mathcal{Y}} p(y \mid \mathbf{x}_q) \log _{2} p(y \mid \mathbf{x}_q),
\end{equation}
as the measure of the total uncertainty. Then, after fixing a set of parameters and considering a conditional distribution of $y$ given the fixed parameters $p(y|\boldsymbol{\theta},\mathbf{x}_q)$, there is no epistemic uncertainty, as the parameters have been settled down. Under such circumstance, the expectation of the entropy of such conditional distribution
\begin{equation}
\mathbf{E}_{p(\boldsymbol{\theta} \mid \mathcal{D})} H[p(y \mid \boldsymbol{\theta}, \mathbf{x}_q)]=-\int p(\boldsymbol{\theta} \mid \mathcal{D})\left(\sum_{y \in \mathcal{Y}} p(y \mid \boldsymbol{\theta}, \mathbf{x}_q) \log _{2} p(y \mid \boldsymbol{\theta}, \mathbf{x}_q)\right) d \boldsymbol{\theta}
\end{equation}
could describe the average aleatoric uncertainty. 

As long as the total and aleatoric uncertainty are obtained, the epistemic uncertainty could be evaluated by their difference 
\begin{equation}
    I(y,\boldsymbol{\theta}) = u_e(\mathbf{x}_q) = H[p(y \mid \mathbf{x}_q)] - \mathbf{E}_{p(\boldsymbol{\theta} \mid \mathcal{D})} H[p(y \mid \boldsymbol{\theta}, \mathbf{x}_q)].
\end{equation}
And this difference equals to the mutual information between $y$ and parameters $\boldsymbol{\theta}$. Intuitively speaking, epistemic uncertainty describes the amount of information about the parameters that could be gained through the true outcome $y$.
\subsection{Ensemble Learning}
\label{ensemble}
Ensemble learning methods leverage multiple machine learning models which produce a series of weak predictive results. For the final output, various voting mechanisms could be applied to fuse all the intermediate results. Ensemble learning has drawn increasing attention and has achieved exceptionally satisfying performance, both in some international machine learning competitions \citep{dong2020survey} and in many application fields, such as financial investment \citep{shen2019kelly} and medical diagnosis \citep{raza2019improving}. Therefore, failure to precisely quantify the uncertainty in these ensemble systems can cause severe consequences.

We start from the classic ensemble model for regression setting. Given observational pairs $\{\mathbf{x},y\}_i \in \mathbb{R}^p \times \mathbb{R}$ which are assumed to be from a data-generating distribution $P^*$, and a series of base model predictors $\{f_k\}_{k=1}^K$ that could be any useful machine learning models, the classic ensemble model assumes the form 
\begin{equation}\label{eq:15}
    y = \sum_{k=1}^K f_k(\mathbf{x})\beta_k + \epsilon,
\end{equation}
where $\boldsymbol{\beta}=\{\beta_k\}_{k=1}^K$ are ensemble weights assigned to each base predictor and $\epsilon$ is a random variable showing the distribution of the outcome.

\cite{DBLP:journals/corr/abs-1911-04061} have introduced a \textit{Bayesian nonparametric ensemble} (BNE) approach to account for different sources of uncertainty for the model in formula \ref{eq:15}. They augment the original model using Bayesian nonparametric machinery and decompose uncertainty on the augmented version of ensemble model. First of all, an extra Gaussian process term $\delta(\mathbf{x})$ is inserted which results in 
\begin{equation}\label{eq:16}
    y = \sum_{k=1}^K f_k(\mathbf{x})\beta_k + \delta(\mathbf{x}) +\epsilon.
\end{equation}
Here $\delta(\mathbf{x})$ has zero mean function $\boldsymbol{0}(x)=0$ and kernel function $k_{delta}(\mathbf{x},\boldsymbol{x'})$; it works as a flexible residual process and adds additional flexibility to the model's mean function. In the case $\epsilon \sim N(0,\sigma_{\epsilon}^2)$, formula \ref{eq:16} could be denoted as a hierarchical Gaussian process $\Phi_{\epsilon}(y|\mathbf{x},\mu)$ where $\mu=\sum_{k=1}^K f_k(\mathbf{x})\beta_k+\delta(\mathbf{x})$ and kernel function is $k_{delta}(\mathbf{x},\mathbf{x'}) +\sigma_{\epsilon}^2$. Next,  $\Phi_{\epsilon}(y|\mathbf{x},\mu)$ is furthered augmented by another Gaussian process $G$ with identity mean function $I(x)=x$, kernel function $k_G$ and a probit-based likelihood constraint. As a result, the final version of the BNE model is 
\begin{equation}\label{eq:17}
    F(y|\mathbf{x},\mu) \sim G(\Phi_{\epsilon}(y|\mathbf{x},\mu), k_G),
\end{equation}
where $\mu=\sum_{k=1}^K f_k(\mathbf{x})\beta_k + \delta(\mathbf{x})$. 

The uncertainty in formula \ref{eq:17} could be decomposed following the paradigm in Figure \ref{fig:branch}. Different choices of the kernel functions of $\delta$ and $G$ consist of the structural uncertainty and training parameters induces parametric uncertainty. Both belong to epistemic uncertainty. 

The entropy $H(y|\mathbf{x})$ is used to measure the overall uncertainty in the predictive distribution $p(y|\mathbf{x})=\int f(y|\mathbf{x}, G, \delta, \boldsymbol{\beta}) dP(G, \delta, \boldsymbol{\beta})$. The aleatoric uncertainty is described by the expected entropy $E_{G, \delta, \boldsymbol{\beta}}[H(y|\mathbf{x},G, \delta, \boldsymbol{\beta})]$ from the model distribution $f(y|\mathbf{x}, G, \delta, \boldsymbol{\beta})$ that is averaged over the posterior belief about $\{G, \delta, \boldsymbol{\beta}\}$. And the epistemic uncertainty is prescribed by the mutual information between $p(G, \delta, \boldsymbol{\beta})$  and $p(y|\mathbf{x})$. Basically, the decomposition of aleatoric and epistemic uncertainty is 
\begin{equation}
    H(y|\mathbf{x}) = \underbrace{I((G, \delta, \boldsymbol{\beta}), y|\mathbf{x})}_{epistemic} + \underbrace{E_{G, \delta, \beta}[H(y|\mathbf{x},G, \delta, \boldsymbol{\beta})]}_{aleatoric}
\end{equation}
which follows the paradigm of \cite{depeweg2018decomposition}. 

Next step is the division of epistemic uncertainty. The parametric uncertainty is about the ensemble weights $\beta_k$ under current specification. Given a specific model setting, such as $\delta=0, G=I$, the parametric uncertainty is encoded in the conditional posterior $p(\boldsymbol{\beta} | \delta=0, G=I)$ and could be measured by the conditional mutual information $I(\boldsymbol{\beta}, y|\mathbf{x}, \delta=0, G=I)$. In other words, parametric uncertainty is the amount of information about parameters $\boldsymbol{\beta}$ we could obtain from the data given such model setting. On the other hand, the structural uncertainty contains two components, uncertainty about the $\delta$ and $G$. The first part describes the uncertainty about $\boldsymbol{\beta},\delta$ under current distribution assumption, i.e. by assuming $G=I$, which is encoded in the difference between distribution $p(\boldsymbol{\beta}, \delta| G=I)$ and $p(\boldsymbol{\beta}| \delta=0, G=I)$. And it could be measured by the change of the mutual information 
\begin{equation}
    I((\boldsymbol{\beta},\delta), y|\mathbf{x}, G=I) - I(\boldsymbol{\beta},y|\mathbf{x} ,\delta=0, G=I).
\end{equation}
Similarly, the second part is the difference betwee $p(\boldsymbol{\beta}, \delta, G)$ and $p(\boldsymbol{\beta}, \delta |G=I)$ and could be measured by the difference of the mutual information. In the end, the overall decomposition of the epistemic uncertainty is 
\begin{equation}
\begin{aligned}
I((\boldsymbol{\beta}, \delta, G), y \mid \mathbf{x}) &=\underbrace{I((\boldsymbol{\beta}, \delta, G), y \mid \mathbf{x})-I((\boldsymbol{\beta}, \delta), y \mid \mathbf{x}, G=I)}_{\text {structural, } G} \\
&+\underbrace{I((\boldsymbol{\beta}, \delta), y \mid \mathbf{x}, G=I)-I(\boldsymbol{\beta}, y \mid \mathbf{x}, \delta=0, G=I)}_{\text {structural }, \delta}+\underbrace{I(\boldsymbol{\beta}, y \mid \mathbf{x}, \delta=0, G=I)}_{\text {parametric }}.
\end{aligned}
\end{equation}

\section{Cross Connections to Other Topics}
\label{connections}
Previous exemplars in Section \ref{decomposition} intend to show a glimpse of uncertainty decomposition in machine learning and they are just a tip of the UQ iceberg. However, these examples have already revealed three mainstream and essential methods, i.e., Gaussian processes, Bayesian inference framework, and ensemble learning.

Section \ref{gaussian} introduced that the variance of the posterior prediction distribution could guide uncertainty decomposition. Based on traditional GP model, \textit{deep Gaussian processes} (DGPs) \citep{damianou2013deep} have been proposed. It could be seen as effective multi-layer extensions of a single GP. Besides, some focus on the choice of kernel functions and propose \textit{deep kernel learning} \citep{AndrewDeepKernel} and \cite{khan2020approximate} investigate the relationship between deep neural networks and GPs.

In Section \ref{bnn}, Bayesian neural networks (BNNs) are introduced to compute uncertainty using entropy of the posterior predictive distribution. Many active sub-fields within BNNs are also essential. \textit{Monte Carlo dropout} \citep{gal2016dropout}, that is, simulation-based inferences, is an effective method to compute the posterior. \textit{Markov chain Monte Carlo} is also an effective method used to approximate inference, some challenges are discussed by \cite{papamarkou2021challenges}. \textit{Variational inference} is another approximation method that learns the posterior distribution over BNNs weights, \cite{davidVI} have provided a thorough review for this field. Last but not least, for a more appropriate selection of model prior, \textit{prior networks} \citep{malinin2018predictive} are proposed which train a deep neural network for an implicit but more flexible prior representation.

Moreover, a decomposition example of ensemble learning is shown in Section \ref{ensemble} where two additional Gaussian process components are added for an accurate estimation. In addition, \cite{FERSINI201426} introduced \textit{Bayesian ensemble learning} which combines ensemble techniques with Bayesian inference. \textit{Deep ensembles} \citep{fort2020deep}, i.e., ensembles of neural networks, are proposed as an alternative to Bayesian neural networks. Besides, \textit{batch ensemble} \citep{wen2020batchensemble} is designed to cost  significantly lower computational and memory resources but still yield competitive accuracy and uncertainties as typical ensembles. 

Most of the methods mentioned above are not just aiming for a decomposition of uncertainty, and hence beyond the scope of this report. However, the way they treat the uncertainty follows the whole paradigm showed in this report. In other words, they still treat the uncertainty as two parts and pursue a more accurate quantification and better mitigation. 

Last but not least, there are other miscellaneous methods that also contribute a lot. \textit{Calibration} \citep{nixon2020measuring} for machine learning models is a hot topic. It makes a model’s predicted probabilities of outcomes reflect true probabilities of those outcomes. \cite{barber2020predictive} introduced \textit{jackknife+}, which is a novel method for constructing predictive confidence intervals. For a more comprehensive review of UQ, readers could refer to \cite{abdar2021review}.

\section{Discussions and Conclusions}
\label{conclusion}

The importance of uncertainty quantification and the necessity to distinguish between various uncertainties, i.e. uncertainty decomposition, have been widely recognized by machine learning researchers. This short report, serving as the first topic in this seminar, has introduced the idea of uncertainty (Section \ref{intro}), clarified two main types of uncertainty (Section \ref{section 1.2}), demonstrated concrete examples of uncertainty decomposition (Section \ref{decomposition}), and showcased relationships with remaining topics (Section \ref{connections}) in this seminar. 

On the other hand, this report has omitted several fields for simplicity. First is the set-based representations of uncertainty, that is, \textit{version space learning}. It can be seen as a counterpart of Bayesian inference, in which predictions are only deterministically described as being possible or impossible rather than probabilities. Despite its limitation, a standpoint worth mentioning is that it is free of aleatoric uncertainty, i.e., all uncertainty is epistemic. Readers could refer to \cite{hullermeier2021aleatoric} for a detailed introduction. 

Secondly, four algorithms mentioned in Section \ref{decomposition} are impossible to cover all domains in machine learning. One of the important missing parts is \textit{reinforcement learning} (RL). In decision-making processes that RL algorithms are trained on, uncertainty still plays a key role and many UQ methods in RL have been proposed. Bayesian techniques are also heavily used in this field. Readers could refer to \cite{zhao2019uncertainty} for a thorough investigation. 

In the end, two conclusions are as follows. Starting with a simplified cat-dog classification task, the necessity of uncertainty quantification and two main types of uncertainty are introduced in Section \ref{intro}. This leads to the first conclusion of this report: \textbf{Uncertainty is certainly critical and we mainly handle two types of uncertainty, i.e., aleatoric uncertainty and epistemic uncertainty.}

Then through the above examples in Section \ref{decomposition}, we could see the pivotal roles of the Bayesian inference framework and Gaussian processes in uncertainty quantification, such as Bayesian neural network in Section \ref{bnn} and the additional Gaussian processes components in Section \ref{ensemble}. Besides, ensemble learning methods also contribute a lot (Section \ref{connections}). Indeed, Bayesian framework, Gaussian processes, and ensemble learning techniques are three of the most widely-used UQ methods in the literature \citep{abdar2021review}. Hence, the second conclusion of this report is that \textbf{there are Bayesian inference frameworks, Gaussian processes, and ensemble models in the arsenal while dealing with uncertainty.}

\section{Acknowledgement}
I would like to thank David Rügamer and Chris Kolb for hosting this seminar and their both detailed and valuable feedback during the preparation. I am also grateful for the suggestions and discussion made by the discussant Faheem Zunjani. 

\newpage

    






    

\RaggedRight
\bibliography{bibliography}
\bibliographystyle{dcu}
\newpage









\end{document}